# Weakly Supervised Cloud Detection Combining Spectral Features and Multi-Scale Deep Network

Shaocong Zhu [a], Zhiwei Li [b], Xinghua Li [c], Huanfeng Shen [a, d, e*]

[a] *School of Resource and Environmental Sciences, Wuhan University, Wuhan 430079, China*

[b] *Department of Civil, Urban, Earth, and Environmental Engineering, Ulsan National Institute of Science and Technology, Ulsan, South Korea*

[c] *School of Remote Sensing and Information Engineering, Wuhan University, Wuhan 430079, China*

[d] *Key Laboratory of Geographic Information System, Ministry of Education, China*

[e] *Key Laboratory of Digital Cartography and Land Information Application, Ministry of Natural Resources, China*

**ABSTRACT:** Clouds significantly affect the quality of optical satellite images, which seriously limits their precise application. Recently, deep learning has been widely applied to cloud detection and has achieved satisfactory results. However, the lack of distinctive features in thin clouds and the low quality of training samples limit the cloud detection accuracy of deep learning methods, leaving space for further improvements. In this paper, we propose a weakly supervised cloud detection method that combines spectral features and multi-scale scene-level deep network (SpecMCD) to obtain highly accurate pixel-level cloud masks. The method first utilizes a progressive training framework with a multi-scale scene-level dataset to train the multi-scale scene-level cloud detection network. Pixel-level cloud probability maps are then obtained by combining the multi-scale probability maps and cloud thickness map based on the characteristics of clouds in dense cloud coverage and large cloud-area coverage images. Finally, adaptive thresholds are generated based on the differentiated regions of the scene-level cloud masks at different scales and combined with distance-weighted optimization to obtain binary cloud masks. Two datasets, WDCD and GF1MS-WHU, comprising a total of 60 Gaofen-1 multispectral (GF1-MS) images, were used to verify the effectiveness of the proposed method. Compared to the other weakly supervised cloud detection methods such as WDCD and WSFNet, the F1-score of the proposed SpecMCD method shows an improvement of over 7.82%, highlighting the superiority and potential of the SpecMCD method for cloud detection under different cloud coverage conditions.

## 1. Introduction

High-resolution optical satellite imagery can be affected by varying degrees of clouds, resulting in different cases of surface information loss. Thick clouds result in a complete loss of information in some areas (Shen et al., 2015), while areas covered by thin clouds can suffer from spectral distortion (Wu et al., 2018). Therefore, many cloud detection techniques have been proposed to improve the usability of optical satellite imagery. The main objective of cloud detection is to identify and segment the cloud region in the image and to provide a mask for the subsequent interpretation and analysis of the image. An accurate cloud mask can minimize the impact of clouds on the subsequent applications of the imagery, such as image reconstruction (Zhu et al., 2023; Yun et al., 2024) and land-cover mapping (Z. Li et al., 2024).

Cloud detection methods based on multi-temporal images (Zhu et al., 2018; Zhang et al., 2021; Liang et al., 2024; Q. Wang et al., 2024; Lee et al., 2025) achieve cloud detection by detecting the abrupt changes in time-series images, but the requirement for two or more images of different time periods limits the practical application of this approach (Zhai et al., 2018). Cloud detection methods based on single images (Ishida et al., 2018; Li et al., 2017; Zhu et al., 2024, 2015) can be categorized into two categories (Wang et al., 2021): 1) physical rule based methods; and 2) machine learning based methods. Given the physical characteristics of clouds, such as the high reflectance and white color (Zhu and Woodcock, 2012), the physical rule based methods tend to achieve cloud detection by designing physical rules and segmentation thresholds. The physical rule based cloud detection methods have a robust performance and high efficiency (Sun et al., 2017). However, the selection of physical rules and thresholds needs to be manually performed, which makes it difficult to obtain the optimal parameters (Z. Wang et al., 2024). This can result in detection leakage and misdetection problems, thereby reducing the usability of the cloud masks.

Machine learning based cloud detection methods train the image classification network with a large-scale dataset to obtain highly accurate cloud detection networks with automated segmentation capabilities. Many of the traditional machine learning methods have been applied to cloud detection tasks, including fuzzy clustering (P. Bo et al., 2020), random forest (Fu et al., 2019; Wei et al., 2020), and support vector machine (Ibrahim et al., 2021; Joshi et al., 2019). Benefiting from the strong feature representation fitting ability of deep learning, deep learning based cloud detection methods trained with pixel-level labels have been widely applied because of the high accuracy that can be achieved. The pixel-level deep learning methods regard cloud detection as an image segmentation task and generate cloud masks via independent prediction on a pixel-by-pixel basis (Li et al., 2019; Chai et al., 2024; J. Li et al., 2024; Wright et al., 2024). Many studies have been conducted to improve the accuracy of cloud detection by designing deep learning networks with different architectures (Chai et al., 2019; Yang et al., 2019; Zhao et al., 2023), introducing multiple image features (Li et al., 2022a; Wang et al., 2023) or incorporating image segmentation techniques (F. Xie et al., 2017; Zi et al., 2018).

However, such methods often require a large amount of well-annotated pixel-level labels to achieve accurate cloud detection. To reduce the workload of manually annotated labels, weakly supervised methods have been proposed. The existing weakly supervised cloud detection methods can be categorized into two methods. The first method utilizes physical rules of clouds to generate pixel-level pseudo-labels for training pixel-level deep learning networks, thereby achieving pixel-level cloud detection (Li et al., 2022b; Liu et al., 2023; Yang et al., 2024; Zhu et al., 2024). The difficulty in defining clear boundaries for thin clouds limits the accuracy of pixel-level pseudo-labels generated by such methods, thereby compromising thin cloud detection performance. The second method regard cloud detection as an image classification task. The images to be detected are segmented into separate small scenes, which are then classified as cloudy or cloudless by the scene-level deep learning networks training with scene-level samples. Compared to pixel-level cloud detection networks, scene-level cloud detection networks exhibit superior generalizability and performance (Shendryk et al., 2019), rendering them more suitable for large-area thin cloud detection tasks. Although some studies have employed a class activation maps (Fu et al., 2018; Li et al., 2020) or a generative adversarial framework (Li et al., 2022b) to generate pixel-level cloud masks from scene-level cloud detection networks. Nevertheless, such methods tend to detect thick clouds with pronounced spectral signatures, which reduces the scene-level network's capability to identify thin clouds.

Although a large number of cloud detection algorithms have been proposed, the existing methods do have some weaknesses: 1) The accuracy of the deep learning based cloud detection methods relies on large-scale, high-quality training samples. However, as shown in Fig. 1, the existing cloud detection datasets (Foga et al., 2017; He et al., 2022; Li et al., 2022a, 2020, 2017; Zhu et al., 2024) often do not cover thin clouds and fog, especially foggy thin clouds. This leads to the fact that most of the current cloud detection methods for high-resolution images can only achieve thick cloud detection. 2) While existing weakly supervised cloud detection methods incorporating spectral features can achieve accurate thick cloud detection, they often fail for thin clouds due to their indistinct features. This limitation persists even when spectral features are integrated into dataset construction or network optimization. Moreover, although spectral features (Kaiming He et al., 2009; Liu et al., 2017) of thin clouds can be leveraged to directly generate relatively high-quality binary masks, the reliance on manual threshold selection and the difficulty in distinguishing clouds from bright surfaces greatly reduces the generality of the methods. 3) Scene-level deep networks are more suitable for large-area thin cloud detection tasks, and single-size scene-level datasets are easy to construct. However, scene-level cloud detection methods based on single-size samples have difficulty in generating pixel-level cloud masks that cover both thick and thin clouds, along with their boundary details.

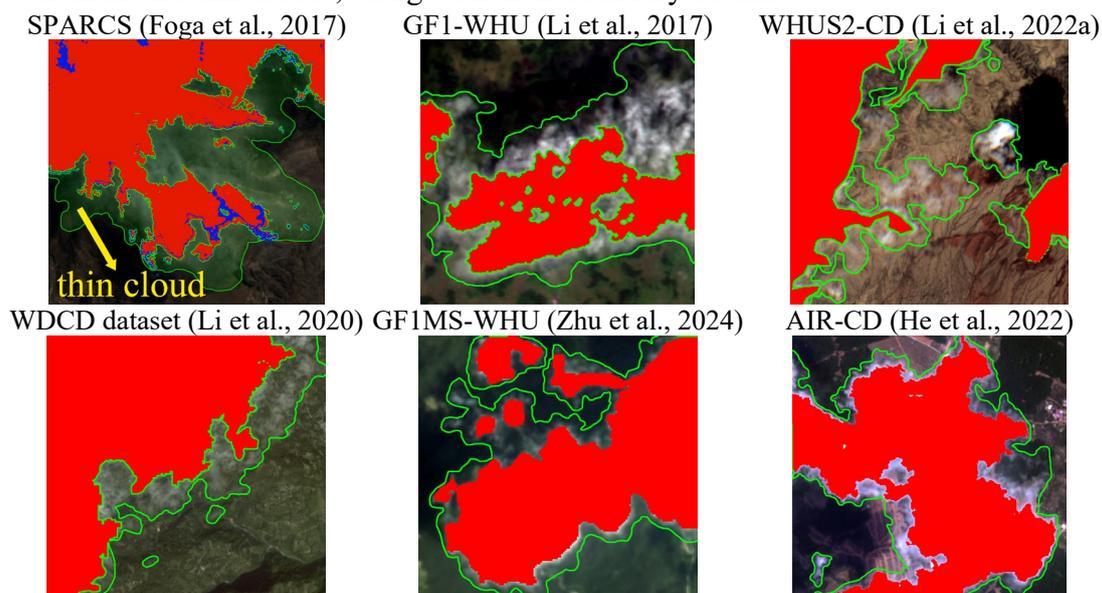

**Fig. 1.** The problem of omitted markers in the existing cloud detection datasets. (Red-color regions denote the original thick label and blue-color regions denote the original thin label.)

To address the challenges of high training data requirements for deep networks, limited thin cloud detection capability, and the single-scale nature of scene-level samples, we propose a weakly supervised cloud detection method that combines cloud thickness map (CTM) and multi-scale scene-level deep network. Specifically, the proposed SpecMCD method constructs multi-scale scene-level labels through sample self-generation and manual supplementation, enabling the training of a multi-scale network to obtain multi-scale cloud probability maps. The cloud probability maps are then combined with CTM to generate thick and thin cloud probability maps according to cloud distribution features across different coverages, followed by fusion based on CTM gradients. A binary cloud mask is automatic segmented using adaptive thresholding and distance weighting. Compared with other weakly supervised methods, SpecMCD achieves effective thin cloud and haze detection by embedding CTM into the inference process, thereby overcoming both the difficulty of detecting thin clouds with ambiguous features and the need for manual threshold selection when incorporating spectral features. In summary, the contributions of this paper are as follows:

1) A weakly supervised learning method for cloud detection is proposed (SpecMCD). By employing the multi-scale scene-level network to suppress bright surfaces and integrating the cloud thickness map (CTM) to enhance thin cloud features, SpecMCD generates cloud probability maps that effectively capture cloud thickness distribution, thereby enabling high-precision pixel-level binary mask extraction.

2) Distinct probability maps are generated for dense and large-area clouds, fused via the CTM gradient, and refined using adaptive thresholding with distance-weighted optimization, enabling automatic and accurate cloud detection.

3) To overcome the limitations of single-scale network, we propose a progressive training framework that integrates multi-scale scene-level samples within a unified network. Furthermore, a local sliding window strategy is adopted to generate multi-scale scene-level cloud probability maps, thereby effectively reducing missed detections.

## 2. Method

The method proposed in this paper consists of three main steps, as shown in Fig. 2: 1) generation of multi-scale scene-level networks and cloud probability maps based on scene-level samples; 2) estimation of the cloud thickness map (CTM) via singular value decomposition; 3) the pixel-level cloud probability map is obtained by combining the multi-scale cloud probability maps and CTM; and 4) extraction of the binary cloud mask using adaptive thresholding combined with distance-weighted optimization.

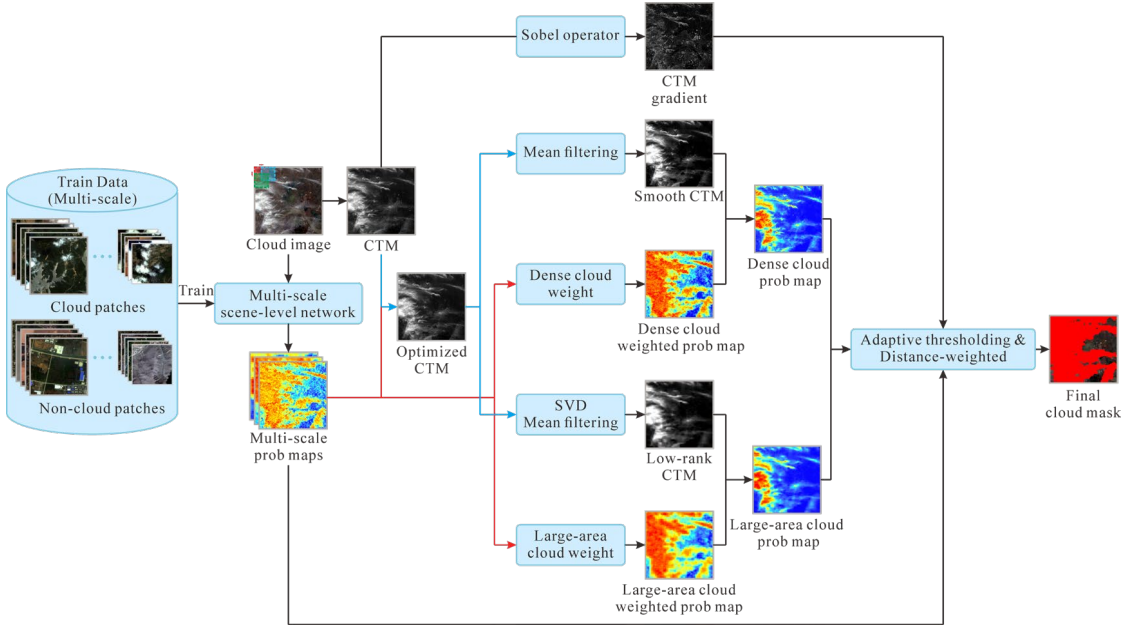

**Fig. 2.** Framework of the proposed weakly supervised cloud detection method.

## 2.1 Generating Multi-Scale Scene-level Network and Cloud Probability Maps Based on Scene-Level Samples

Most existing scene-level datasets contain samples of a single size, which limits the ability of cloud detection networks to accurately detect clouds with varying coverage. Large-scale cloud detection network are prone to suffering from the loss of usable information, while small-scale cloud detection network suffer from missed detections. Therefore, this study utilizes multi-scale scene-level samples to train multi-scale scene-level network. The resulting dataset includes images at three resolutions (256×256, 128×128, and 64×64) and comprises thick cloud, thin cloud, and cloud-free samples. Thin cloud sample were obtained by manually outlining rough cloud masks for images containing large areas thin clouds, whereas thick cloud samples were generated using the scene-level pseudo-label generation strategy from the TransMCD method (Zhu et al., 2024).

In this study, the multi-scale scene-level dataset was utilized to train the RegNetY network (Radosavovic et al., 2020), which comprises three main parts, as shown in Fig. 3: 1) a stem incorporates a stride two 3 × 3 convolutional layer with 32 output channels; 2) a body consists of multiple downsampled stages. Each stage contains a series of blocks, which are composed of standard residual bottleneck blocks with group convolution (S. Xie et al., 2017) and Squeeze-and-Excitation (Hu et al., 2020) attention mechanism; 3) a head component contains average pooling, fully connected and dropout layers (Cao and Huang, 2022). During training, each image block was assigned a single binary label indicating the presence or absence of clouds, and the network output was configured with two channels accordingly.

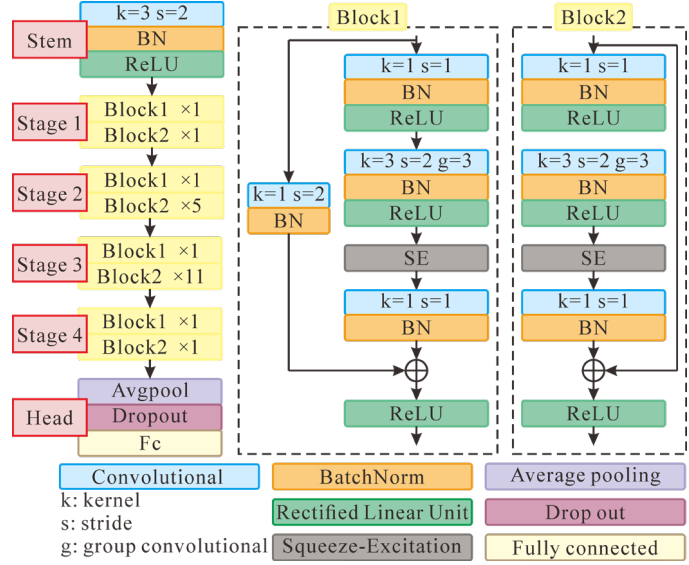

**Fig. 3.** Structure of the RegNetY-040 network.

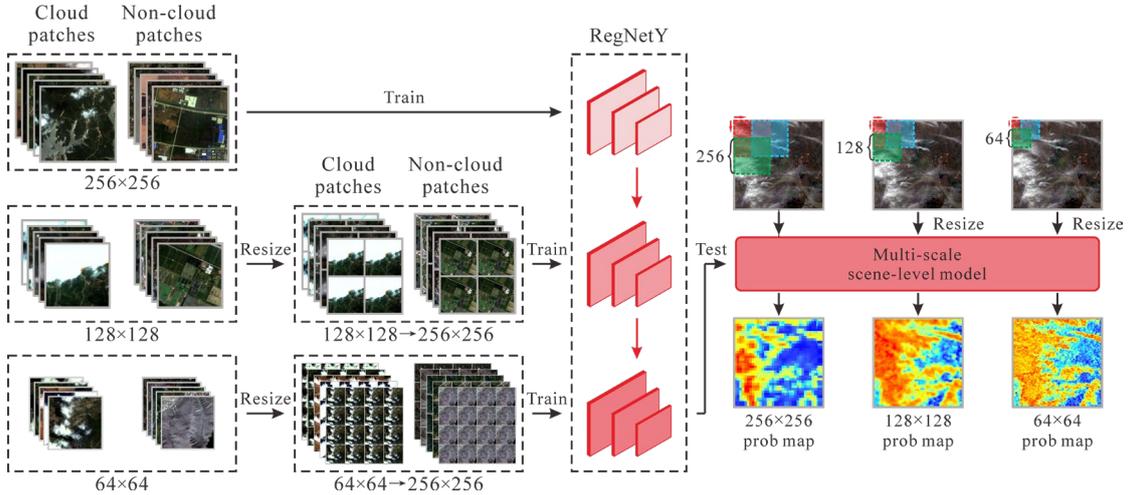

**Fig. 4.** The progressive training framework for generating multi-scale scene-level network and cloud probability maps.

To fully leverage scene-level samples at different scales and generate multi-scale scene-level cloud probability maps using a single network, a progressive training framework is adopted, as illustrated in Fig. 4. Firstly, the network scale is set to 256 × 256. Each 128 × 128 sample is replicated into 4 copies and each 64 × 64 sample is replicated into 16 copies to resize the multi-scale samples to match the target size. Since small-scale samples contain fewer thin cloud features, making them challenging for effective detection, the network is first trained with large-scale samples, and small-scale samples are gradually incorporated to enhance the network's capability in generating multi-scale scene-level cloud probability maps that capture thin clouds. Multi-scale probability maps are then generated using a local sliding window strategy (Li et al., 2020) to reduce missed detections. Window sizes are set to 256 × 256, 128 × 128, and 64 × 64 according to the training sample sizes. Overlapping image blocks are generated by sliding the window from left to right and top to bottom with a step size of half the window size. All blocks are resized to 256 × 256 using the same progressive resizing strategy. The scene-level network classifies each block individually, and the predicted probability is assigned to the corresponding block. For overlapping regions, the maximum probability among overlapping blocks is taken as the final value to achieve block-level fusion. Finally, a scene-level binary cloud mask is generated by applying an initial threshold of 0, with all regions exceeding this threshold classified as cloud-covered.

**2.2 Generating CTM based on singular value decomposition**

Since the blue band is most strongly affected by thin clouds, an initialized CTM capturing spectral features of clouds can be esstimated by a synthetic band (Liu et al., 2017). The initialized CTM is calculated as follows:

$$CTM = 2 * B - 0.95 * G \qquad (1)$$

where $B$ and $G$ are the values of the image blue band and green band, respectively.

However, the initialized CTM obtained by the $B$ and $G$ bands is affected by highlighted surfaces, resulting in both cloud-covered areas and bright surfaces to appear as high-intensity regions. Therefore, regions with CTM values exceeding the median are identified as highlighted surfaces. Highlighted cloud-free regions are then obtained by comparing these surfaces with the intersections of multi-scale scene-level cloud masks. is subsequently refined by reducing the CTM values in the highlighted cloud-free regions to half of their original value, thereby mitigating the impact of bright surfaces on the initialized CTM.

Although the initialized CTM effectively highlight the spectral features of clouds, the distribution characteristics differ considerably between dense cloud and large area cloud images. Clouds in dense cloud images are characterized by dispersed and irregular arrangements, while clouds in large-area cloud images are characterized by wide coverage, and the cloud boundaries are difficult to identify. Therefore, differentiated CTM optimization strategies were applied to enhance the spectral features of both dense and large-area clouds.

For dense cloud images, the CTM is smoothed using mean filtering to suppress noise. For large-area cloud images, where local details may reduce the global cloud features, the CTM is decomposed into two orthogonal matrices and one diagonal matrix by singular value decomposition (SVD). A low-rank approximation is then obtained by retaining the first 70 singular values, as described in (2):

$$CTM_{SVD} = U_k \Sigma_k V_k^T \qquad (2)$$

where $CTM_{SVD}$ is the low-rank approximation of the CTM, $U_k$ is the first $k$ columns of the left singular value matrix, $\Sigma_k$ is the first $k$ singular values, and $V_k^T$ is the transpose of the first $k$ rows of the right singular matrix, which $k$ default to 70.

**2.3 Generating Pixel-Level Cloud Probability Maps by Combining CTM and Multi-Scale Cloud Probability Maps**

Scene-level networks are constrained by scene size and often misclassify cloud-free regions between dispersed cloud blocks as cloudy, resulting in a significant loss of usable information in dense cloud images. Furthermore, the indistinct spectral features of thin clouds hinder deep learning networks from achieving accurate detection, leading to a large number of omissions in large-area cloud images. To address these issues, the proposed method integrates multi-scale cloud probability maps with CTM to generate pixel-level cloud probability maps based on the characteristics of both dense and large-area cloud images.

### 2.3.1 Generating the Large-Area Cloud Probability Map

To improve the ability of the proposed method to detect thin clouds in large-area cloud images, we first construct a large-area cloud-weighted probability map by aggregating the multi-scale cloud probability maps using large-area cloud weight coefficients. This weighted probability map is then multiplied by the low-rank CTM to obtain the large-area cloud probability map. The probability of each pixel in the large-area cloud probability map is calculated as follows:

$$\rho_{Large(i,j)} = (\mu_1 \cdot \rho_{256(i,j)} + \mu_2 \cdot \rho_{128(i,j)} + \mu_3 \cdot \rho_{64(i,j)}) \cdot CTM_{SVD(i,j)} \quad (3)$$

where $\rho_{Large(i,j)}$ is the large-area cloud probability of a pixel at row $i$, column $j$ of the image; $\rho_{256}$, $\rho_{128}$, and $\rho_{64}$ are the normalized cloud probability obtained by the three different-scale scene-level networks of 256 × 256, 128 × 128, and 64 × 64, respectively; $CTM_{SVD}$ is the normalized low-rank CTM; $\mu_1$, $\mu_2$, and $\mu_3$ are constants, which default to 0.5, 0.4, and 0.1, respectively.

### 2.3.2 Generating the Dense Cloud Probability Map

To address the tendency of scene-level networks to misdetect in dense cloud images, the proposed method aggregates the multi-scale cloud probability maps using dense cloud weight coefficients to obtain a dense cloud weighted probability map. This weighted map is then multiplied by the smoothed CTM to generate the dense cloud probability map. The probability of each pixel in the dense cloud probability map is calculated as follows:

$$\rho_{Dense(i,j)} = (\mu_3 \cdot \rho_{256(i,j)} + \mu_2 \cdot \rho_{128(i,j)} + \mu_1 \cdot \rho_{64(i,j)}) \cdot CTM_{Mean(i,j)} \quad (4)$$

where $\rho_{Dense(i,j)}$ is the dense cloud probability of the pixel at row $i$, column $j$ of the image; $CTM_{Mean}$ is the normalized CTM smoothed by the mean filtering; and the mean filtering window size is set to 29.

### 2.3.3 Fusing the Dense Cloud and Large-Area Cloud Probability Maps

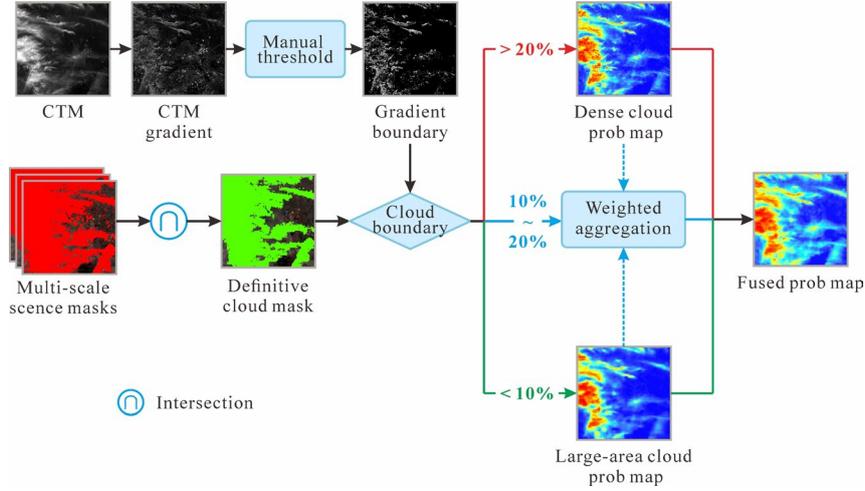

**Fig. 5.** Flowchart for fusing the dense cloud and large-area cloud probability maps.

Since the proposed method generates two different types of cloud probability maps, manually determining the cloud type for each image remains time-consuming and labor-intensive. Thick clouds in the image have distinct boundaries, whereas thin clouds lack clear boundaries. Correspondingly, the CTM gradient is higher at thick cloud boundaries and lower at thin cloud boundaries. Leveraging these characteristics, we propose a fusion strategy for dense and large-area cloud probability maps based on the CTM gradient, as illustrated in Fig. 5. Specifically, the Sobel operator is applied to compute the CTM gradient and generate the binary gradient boundary mask as follows:

$$M_{Bound}(i,j) = \begin{cases} 1, if\ Grad(i,j) > \mu_{Grad} \\ 0, else \end{cases} \quad (5)$$

where $M_{Bound}$ is the binary gradient boundary mask; $Grad(i,j)$ is the normalized CTM gradient of the pixel at row $i$, column $j$ of the image; $\mu_{Grad}$ is a constant, which defaults to 19.

By calculating the proportion of $M_{Bound}$ within the intersections of the multi-scale cloud masks, the dense cloud and large-area cloud probability maps are fused as follows:

$$\rho_{Fused(i,j)} = \begin{cases} \rho_{Dense(i,j)}, if\ P \geq \mu_1 \\ k \cdot \rho_{Dense(i,j)} + (1-k) \cdot \rho_{Large(i,j)}, if\ \mu_2 < P < \mu_1 \\ \rho_{Large(i,j)}, if\ P \leq \mu_2 \end{cases} \quad (6)$$

where $\rho_{Fused(i,j)}$ is the fused cloud probability of the pixel at row $i$, column $j$ of the image; $P$ is the percentage of the $M_{Bound}$ coverage area within $M_{Cloud}$; $k$ is the fusion constant, calculated as $(P - \mu_2)/(\mu_1 - \mu_2)$, $\mu_1$ and $\mu_2$ are empirical constants, which default to 0.2 and 0.1, respectively.

### 2.4 Generating Pixel-Level Cloud Probability Maps by Combining CTM and Multi-Scale Cloud Probability Maps

To eliminate the need for manual threshold adjustment and enhance the robustness of the method, adaptive segmentation thresholds are automatically extracted by analyzing the differences in regions detected by the scene-level network across scales. The thresholds for different images are calculated as follows:

$$\mu_{Final} = \begin{cases} \mu_{Dense}, if\ P \geq \mu_1 \\ \cdot \mu_{Dense} + (1-k) \cdot \mu_{Large}, if\ \mu_2 < P < \mu_1 \\ \mu_{Large}, if\ P \leq \mu_2 \end{cases} \quad (7)$$

where $\mu_{Final}$ denotes adaptive threshold; $\mu_{Dense}$ is the dense cloud threshold, obtained by averaging the probabilities within the region covered by the 256 × 256 scene-level cloud mask; and $\mu_{Large}$ represents the large-area cloud threshold, obtained by averaging the probabilities of the regions differing between the 256 × 256 and 64 × 64 scene-level cloud masks.

Based on $\mu_{Final}$, the fused cloud probability map is segmented and the initialized binary cloud mask $M_{Init}$ is generated as follows:

$$M_{Init(i,j)} = \begin{cases} 1, if\ \rho_{Fused(i,j)} > \mu_{Final} \\ 0, else \end{cases} \quad (8)$$

$M_{Init}$ is derived from $\mu_{Final}$ rather than manual thresholding, making it difficult to accurately detect thin clouds. Therefore, an adaptive expansion strategy is introduced by augmenting $\rho_{Fused}$ around $M_{Init}$ through distance-weighted optimization to enhance the thin cloud detection capability of the SpecMCD method, as shown in (9). Finally, the final binary cloud mask $M_{Final}$ is obtained by segmenting the distance-weighted cloud probability map using $\mu_{Final}$.

$$\rho_{Dist(i,j)} = \rho_{Fused(i,j)} + \frac{DistMax - Dist}{DistMax} \cdot \rho_{Mean} \quad (9)$$

where $\rho_{Dist}$ is the cloud probability map after the distance-weighted optimization; $DistMax$ is the adaptive distance constant, decreasing from 100 to 50, depending on $P$, calculated as $(150 - P * 500)$; $Dist$ is the distance from the pixel at row $i$, column $j$ to the nearest cloud block; and $\rho_{Mean}$ is the compensation probability, which is obtained by averaging the $\rho_{Fused(i,j)}$ of the region covered by the 128 × 128 scene-level cloud mask.

## 3. Experiment data and result

### 3.1 Experiment data and setting

Since a single-scale scene-level network cannot adequately capture cloud coverage across different distribution patterns, it is necessary to introduce multi-scale scene-level labels to train multi-scale networks for cloud detection. In this study, 101 cloud images and 114 cloud-free images were used to generate multi-scale thick-cloud and cloud-free scene-level samples using the scene-level pseudo-label generation strategy from the TransMCD method (Zhu et al., 2024). To address the difficulty of automatically generating thin-cloud samples, 10 images with large-area thin clouds were roughly annotated to provide approximate thin-cloud regions, which were used exclusively for constructing thin cloud scene-level samples, as shown in Fig. 6. Based on these data, a multi-scale scene-level training dataset was constructed, comprising 21 699 image blocks of size 256 × 256, 84 845 image blocks of size 128 × 128, and 332 251 image blocks of size 64 × 64, for training the multiple scene-level network, as shown in Table 1. The bandwidth information for each band in the visible light band is as follows: blue (0.45–0.52 μm), green (0.52–0.59 μm), and red (0.63–0.69 μm).

**Table 1**
Summary of the experiment data utilized in this study.

| Dataset source | Image source | Image size | Number of images | Label type | Usage |
| --- | --- | --- | --- | --- | --- |
| Multi-scale Dataset | GF1-MS | 256 × 256 | 61,336 | Scene-level | Train multi-scale scene-level network |
| | | 128 × 128 | 224,885 | | |
| | | 64 × 64 | 826,705 | | |
| WDCD & GF1MS-WHU | GF1-MS | 256 × 256 | 13,608 | Pixel-level | Train pixel-level network |
| | | > 4500 × 4500 | 18 | Pixel-level | Validation |

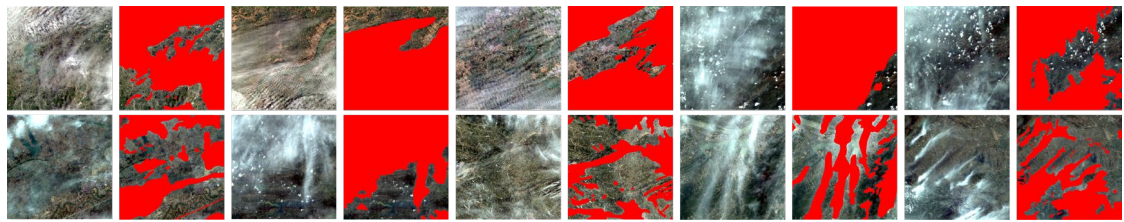

**Fig. 6.** The large-area thin cloud images in the training dataset with multi-scale scene-level labels. Each image is followed by its manual labels, where the red-color regions denote the cloud regions.

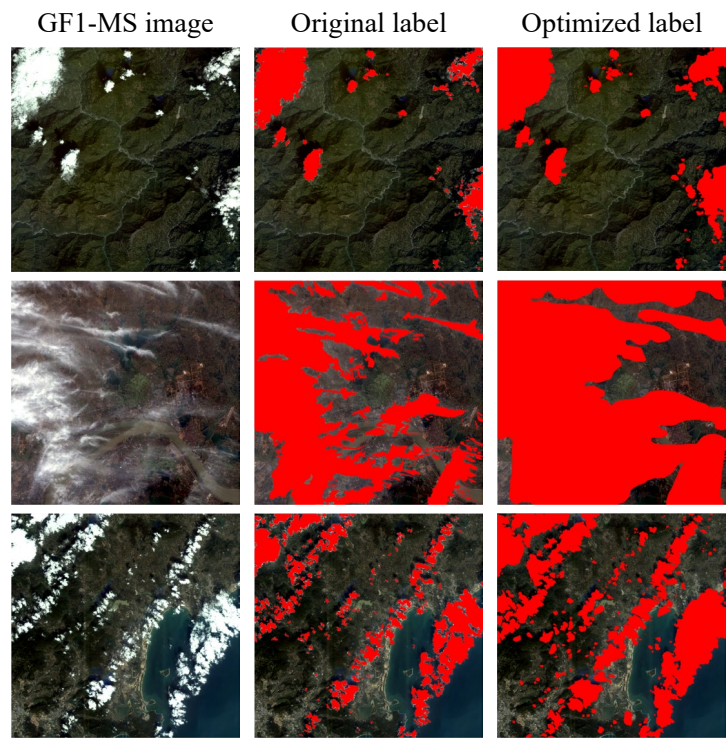

**Fig. 7.** Visual examples of the original and optimized labels.

To comprehensively evaluate the performance of different cloud detection methods, the validation and test sets in the WDCD (Li et al., 2020) dataset and the GF1MS-WHU (Zhu et al., 2024) dataset were combined to generate a new dataset containing 60 GF1-MS images. Considering the issue of missing annotations in the original pixel-level labels, we refined the labels to enhance their accuracy for thin clouds, with examples shown in Fig. 7. The refined labels provide a more accurate representation of cloud coverage, although minor misclassifications remain. Overall, the omission problem was effectively alleviated, making the refined labels more suitable for subsequent applications. Finally, this dataset with pixel-level labels was randomly divided at a 7:3 ratio and used as training data for fully supervised deep learning methods, as well as validation data for all cloud detection methods.

In this study, all RegNetY networks were implemented using the pre-trained RegNetY_040 (Radosavovic et al., 2020) network from the timm library (Wightman, 2019) and optimized with the Adam optimizer. The hyper-parameters were configured as follows: $\beta_1$ was set to 0.9, and $\beta_2$ was set to 0.999, and the loss function was cross-entropy with softmax. Training was conducted for 100 epochs with an initial learning rate of $1\times10^{-4}$, which was reduced by a factor of 0.1 after 25 epochs. In the progressive training framework, 128 × 128 and 64 × 64 samples were incorporated at the 30th and 60th epochs, respectively. The hyper-parameters of other compared methods were set according to their original publications. Data normalization was performed by dividing each pixel by the maximum value within its corresponding image block. All networks were trained on a Windows equipped with an Intel Core i7-10700 CPU @ 2.90 GHz, 32 GB RAM, and an NVIDIA GeForce RTX 3070Ti GPU with 8 GB of memory.

## 3.2 Comparison With Weakly Supervised Cloud Detection Methods

**Table 2**

Quantitative evaluation of the binary cloud masks obtained by the weakly supervised cloud detection methods.

| Method | Supervision | OA | Precision | Recall | F1-score | F2-score |
|---|---|---|---|---|---|---|
| HCDNet | Rule | 0.6847 | 0.9529 | 0.4474 | 0.5836 | 0.4916 |
| TransMCD | | 0.6868 | **0.9998** | 0.4577 | 0.6049 | 0.5068 |
| SL-256 | | 0.7454 | 0.6393 | **0.9715** | 0.7330 | 0.8404 |
| SL-128 | | 0.8051 | 0.6960 | 0.9635 | 0.7807 | 0.8695 |
| SL-64 | Weakly supervised | 0.8429 | 0.7464 | 0.9542 | 0.8215 | 0.8906 |
| WSFNet | | 0.6787 | 0.7439 | 0.6231 | 0.6119 | 0.6086 |
| WDCD | | 0.8028 | 0.7848 | 0.8599 | 0.7847 | 0.8185 |
| SpecMCD | | **0.9126** | 0.8815 | 0.9287 | **0.8997** | **0.9156** |

To validate the effectiveness of the proposed SpecMCD method, we conducted a comparative analysis against weakly supervised cloud detection methods, including HCDNet (Liu et al., 2023), TransMCD (Zhu et al., 2024), WSFNet (Fu et al., 2018), WDCD (Li et al., 2020) and three baseline scene-level RegNetY (Radosavovic et al.,

2020) networks at different scales, namely, SL-256, SL-128, and SL-64. Among them, WSFNet, WDCD, and the baseline RegNetY networks rely solely on scene-level labels, whereas HCDNet and TransMCD leverage physical rules to generate pseudo-labels to improve cloud detection accuracy. The performance of all methods was evaluated using overall accuracy (OA), Precision, Recall, F1-score, and F2-score. Table 2 demonstrate that the physics rule-based HCDNet and TransMCD methods fail to generate reliable pseudo-labels for thin clouds, resulting in substantial under-detection of thin cloud regions, with the recall lower than 0.46. The baseline scene-level networks, SL-256, SL-128, SL-64, can effectively avoid the problem of detection leakage, achieving a recall exceeding 0.95 in all the binary cloud masks. However, the baseline scene-level networks tend to exhibit significant misdetection, with the precision lower than 0.75. And as the scale decreases, the misdetection occurrences of the scene-level networks are improved, but the detection leakage deteriorates, resulting in low OA, F1-score, and F2-score values for the baseline scene-level networks at different scales. The WSFNet and WDCD method employ a class activation mapping-like mechanism to refine scene-level network for generating pixel-level binary cloud masks. However, this refinement process leads to a reduction in Recall compared to the original scene-level networks. Compared with other weakly supervised methods, the SpecMCD method refines the scene-level cloud masks based on the spectral feature of clouds, achieving significant improvements of over 6.97%, 7.82%, and 2.50% in OA, F1-score and F2-score, respectively.

To further assess the performance of different weakly supervised cloud detection methods, a visual comparison of the binary cloud masks was performed, as shown in Fig. 8. The results show that physics rule-based HCDNet and TransMCD methods are effective in thick cloud detection but exhibit severely limited capability in detecting thin clouds under both dense and large-area cloud cover. Baseline scene-level networks at different scales demonstrate certain advantages in large-area cloud coverage regions but misclassify cloud-free regions between dense cloud blocks, leading to significant errors. WSFNet, which emphasizes thin cloud features during training, tends to neglect thick cloud structures, resulting in major omission errors in dense cloud images. WDCD improves precision in dense cloud images through its class attention mechanism but remains vulnerable to minor misdetections and severe detection leakage in large-area cloud images. By contrast, the proposed SpecMCD method applies differentiated processing strategies for varying cloud coverage scenarios. Visual results confirm its superior performance in large-area cloud regions, enabling more comprehensive cloud detection. In dense cloud scenes, SpecMCD also demonstrates superior results, though performance near urban areas with high CTM values may be similar to that of SL-64.

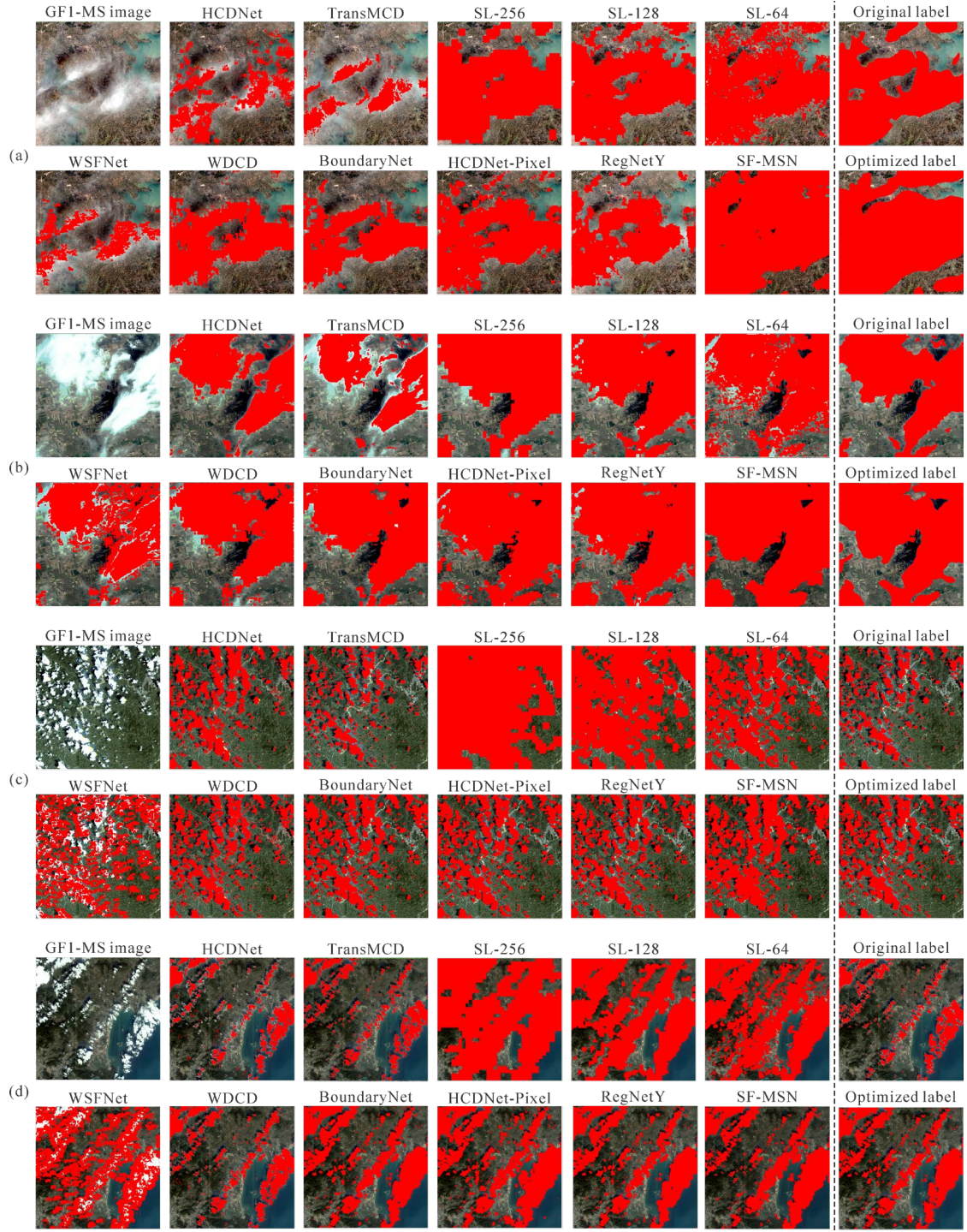

**Fig. 8.** Examples of binary cloud masks obtained by weakly and fully supervised cloud detection methods.

### 3.3 Comparison With Full Supervised Cloud Detection Methods

In this section, we compare the proposed weakly supervised SpecMCD method with state-of-the-art fully supervised methods, including BoundaryNet (Zhao et al., 2023), HCDNet-Pixel (Liu et al., 2023) and RegNetY (Radosavovic et al., 2020). A quantitative evaluation is presented in Table 3. The results show that the fully supervised methods achieve high accuracy, with F1-scores exceeding 0.88, and benefit from strong feature extraction capabilities that reduce misdetection. However, they still struggle with thin clouds and haze due to their indistinct spectral features. In contrast, SpecMCD achieves superior performance, improving overall accuracy (OA) and F2-score by more than 1.59% and 1.40%, respectively, compared with the fully supervised methods.

Fig. 8 also presents the binary cloud masks obtained by BoundaryNet, HCDNet-Pixel, RegNetY, and SpecMCD. From the visualization results, it can be seen that the BoundaryNet method over-emphasizes boundary details of cloud, which enables it to generate relatively precise cloud masks in dense cloud regions. However, it is not suitable for identifying large-area cloud regions. HCDNet-Pixel exhibits stronger thin cloud detection than other fully supervised methods, though it introduces minor misdetections. RegNetY achieves more balanced overall performance. Despite training with optimized pixel-level labels, the fully supervised methods still struggle with accurate detection in large-area cloud regions, particularly for thin clouds. In contrast, SpecMCD demonstrates superior performance in large-area cloud detection, enabling more comprehensive cloud coverage identification. However, in dense cloud regions, SpecMCD shows some misdetection and its ability to capture fine cloud details remains weaker than that of the fully supervised methods.

**Table 3**

Quantitative evaluation of the binary cloud masks obtained by the full supervised cloud detection methods.

| Method | Supervision | OA | Precision | Recall | F1-score | F2-score |
| --- | --- | --- | --- | --- | --- | --- |
| BoundaryNet | Full supervised | 0.8734 | 0.9221 | 0.8689 | 0.8813 | 0.8717 |
| HCDNet-Pixel |  | 0.8878 | 0.8842 | 0.9126 | 0.8899 | 0.9016 |
| RegNetY |  | 0.8967 | **0.9461** | 0.8781 | **0.9029** | 0.8866 |
| SpecMCD | Weakly supervised | **0.9126** | 0.8815 | **0.9287** | 0.8997 | **0.9156** |

## 4. Discussion

### 4.1 Effectiveness of the Progressive Training Framework

To verify the effectiveness of the proposed progressive training framework, we conducted comparative experiments between binary masks obtained by the multi-scale scene-level network trained using the progressive framework (denoted as SL-Stack-256, SL-Stack-128, and SL-Stack-64) and those generated by baseline scene-level networks trained solely on single-scale samples. In addition, we evaluated the accuracy of pixel-level binary masks obtained from both the progressively trained multi-scale network and the single-scale baseline networks (SpecMCD-Base) within the proposed method, as shown in Table 4. The results show that the progressive framework consistently improves OA, Recall, F1-score, and F2-score across cloud masks at different scales, compared to single-scale training. These improvements confirm that the proposed progressive training framework effectively enhances the ability of the scene-level network to obtain accurate multi-scale cloud masks. Furthermore, comparative experiments between SpecMCD-Base and SpecMCD demonstrate that incorporating high-precision multi-scale scene-level network further improves the accuracy of pixel-level cloud masks within the proposed method.

**Table 4**

Quantitative evaluation of the binary cloud masks obtained by the full supervised cloud detection methods.

| Method | OA | Precision | Recall | F1-score | F2-score |
| --- | --- | --- | --- | --- | --- |
| SL-256 | 0.7454 | 0.6393 | 0.9715 | 0.7330 | 0.8404 |
| SL-128 | 0.8051 | 0.6960 | 0.9635 | 0.7807 | 0.8695 |
| SL-64 | 0.8429 | 0.7464 | 0.9542 | 0.8215 | 0.8906 |
| SpecMCD-Base | 0.9040 | 0.8702 | 0.9309 | 0.8945 | 0.9147 |
| SL-Stack-256 | 0.7505 | 0.6287 | **0.9913** | 0.7375 | 0.8540 |
| SL-Stack-128 | 0.8282 | 0.7038 | 0.9815 | 0.7963 | 0.8878 |
| SL-Stack-64 | 0.8822 | 0.7848 | 0.9674 | 0.8536 | 0.9139 |
| SpecMCD | **0.9126** | **0.8815** | 0.9287 | **0.8997** | **0.9156** |

## 4.2 Sensitivity Analysis of the Hyperparameters

The SpecMCD method introduces several hyperparameters for generating pixel-level cloud probability maps after incorporating spectral feature. To validate the robustness of the proposed method, we conducted a sensitivity analysis of the mean filtering window size, the selected singular values in SVD decomposition and the threshold for obtaining the binary gradient boundary mask.

Fig. 9(a) illustrates the impact of mean filtering window size on smoothing CTM in dense cloud images and its influence on the accuracy of the final binary cloud mask. Fig. 10 further compares the dense cloud probability maps and binary cloud masks before and after mean filtering. The results demonstrate that mean filtering tends to distort dense cloud shapes, causing the omission of fine-scale details in the binary masks and leading to a decline in the F2-score. Nevertheless, an appropriately sized mean filter can effectively suppress noise and reduce false positives from small bright non-cloud surfaces, thereby improving OA and F1-scores and yielding more balanced detection performance.

As shown in Fig. 9(b), when extracting cloud features from large-area cloud images using SVD decomposition, selecting singular values greater than 8 has little effect on the final cloud mask. Furthermore, Fig. 9 (c) presents a sensitivity analysis of the threshold $\mu_{Grad}$ for obtaining the binary gradient boundary mask. Since CTM gradient tends to be higher at thick cloud boundaries, an excessively small $\mu_{Grad}$ may cause boundary extraction errors, hereby reducing the accuracy and stability of the final mask. When $\mu_{Grad}$ is set within the range of 17–21, its influence on the accuracy of the binary cloud mask is minimal, indicating that SpecMCD is robust to this hyperparameter within a reasonable range. However, setting $\mu_{Grad}$ beyond this range leads to degraded thick cloud boundary detection and consequently lower mask accuracy.

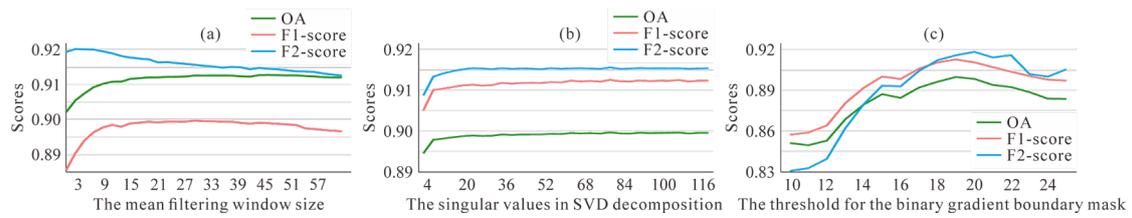

**Fig. 9.** Sensitivity analysis of hyperparameters in generating pixel-level cloud probability maps. (a) The mean filtering window size for smoothing CTM in dense cloud images. (b) The singular values $k$ in SVD decomposition. (c) The threshold $\mu_{Grad}$ for obtaining the binary gradient boundary mask.

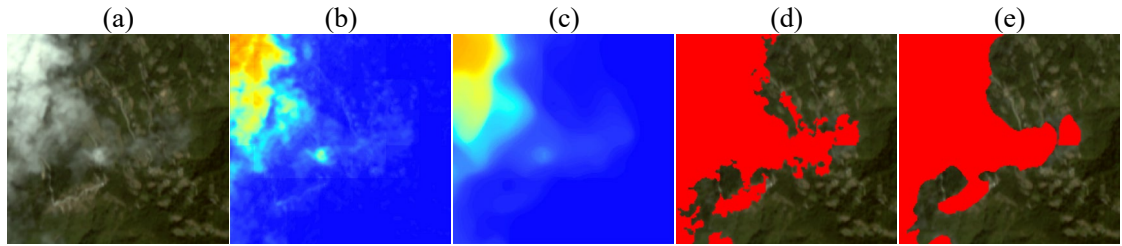

**Fig. 10.** Visual examples of dense cloud probability maps and binary cloud masks before and after mean filtering. (a) Original GF1-MS image. (b) Original cloud probability map without mean filtering. (c) Mean filtered cloud probability map with 29 window size. (d) Original binary cloud mask without mean filtering. (e) Mean filtered binary cloud mask with 29 window size.

### 4.3 Analysis of the Cloud Probability Maps

Spectral feature-based methods can achieve satisfactory performance in certain images (Foga et al., 2017). However, they require manual threshold adjustments across different sensors or even across scenes from the same sensor, which limits their robustness and automation compared with machine learning–based methods. While the proposed SpecMCD method incorporates spectral features, this integration may reduce the automatic segmentation capability of deep learning networks. Therefore, we employ adaptive thresholding to generate the binary cloud mask, rather than an artificially optimal threshold. Nevertheless, this mask still struggles to accurately delineate thin clouds surrounding thick cloud regions. Even with distance weighting optimization, the absence of a fixed scope for thin clouds leads to misdetections in dense cloud areas and detection leakage in large-area cloud regions.

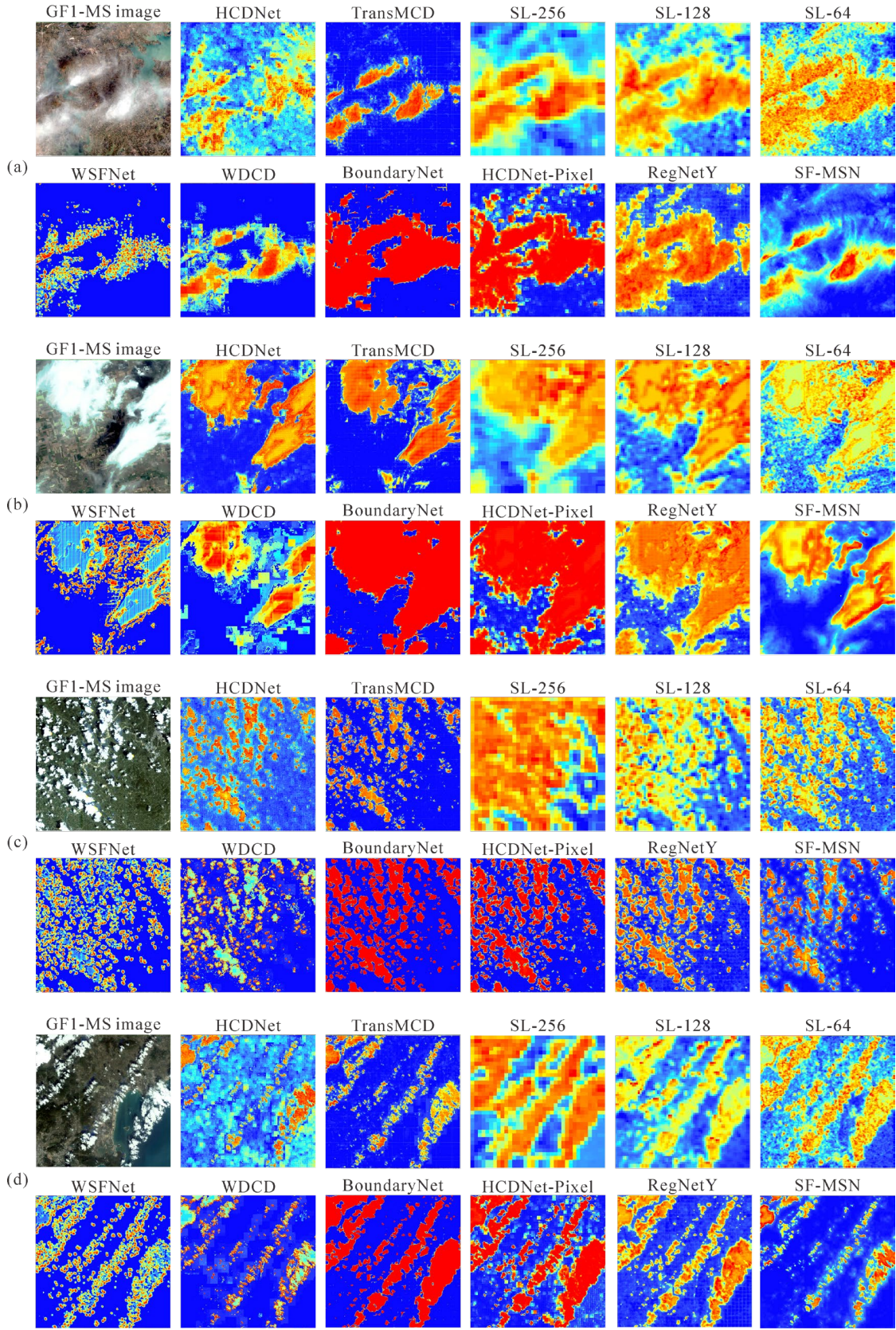

**Fig. 11.** Examples of cloud probability maps obtained by weakly and fully supervised cloud detection methods. The cloud probability maps depict the color scale from blue (low cloud probability) to red (high cloud probability).

To further investigate this limitation, we compared cloud probability maps generated by different methods without binarization. Fig. 11 shows the cloud probability maps from HCDNet, TransMCD, WSFNet, and WDCD effectively capture thick cloud distributions but show limited representation of thin clouds, hindering accurate binary mask generation. The baseline scene-level networks display strong capability in representing cloud probabilities over large areas; however, their ability diminishes as the network scale decreases. Fully supervised methods, including BoundaryNet, HCDNet-Pixel, and RegNetY, achieve reliable cloud probability estimation in dense cloud regions. However, despite being trained with thin cloud samples, they fail to capture the gradual transition from thick to thin clouds, producing probability maps that lean toward binarization. The SpecMCD method generates cloud probability maps in dense cloud regions that surpass those of other weakly supervised methods and approach the quality of RegNetY. Nevertheless, SpecMCD still loses fine-scale details, particularly at cloud boundaries, resulting in slightly inferior performance compared with fully supervised methods. In large-area cloud images, SpecMCD more effectively represents cloud thickness and can better represent the cloud distribution in the image.

## 4.4 Limitations and Future Perspectives

The SpecMCD method can generate pixel-level cloud masks using scene-level labels. While it achieves superior cloud detection in large-area images compared to fully supervised methods, its performance in dense cloud images remains limited. Since distinguishing clouds from snow/ice using only visible and near-infrared bands is inherently challenging (Li et al., 2022a), the reliance of SpecMCD on visible bands for spectral feature extraction makes it heavily dependent on the multi-scale scene-level network. As shown in Fig. 12, misclassification of snow as cloud occurs when the multi-scale network fails to correctly differentiate the two.

Furthermore, based on the FRARC (Zhu et al., 2023) thick cloud removal method, we evaluated the cloud reconstruction performance using different cloud masks, as illustrated in Fig. 13. Although RegNetY-based reconstruction can effectively remove thick clouds, it remains sensitive to undetected thin clouds, often resulting in radiance overestimation. In contrast, SpecMCD-based reconstruction benefits from thin cloud detection, enabling simultaneous removal of thick and thin clouds and obtaining visually satisfactory results. However, for regions with particularly thin cloud cover, SpecMCD still encounters challenges in achieving accurate detection, such as the region marked by the blue box in Fig. 13(c).

In future work, we aim to combine pixel-level and scene-level networks to improve the cloud detection accuracy in dense cloud regions. To address misclassification between clouds and snow, we will incorporate manually labeled cloud–snow samples in a data-driven framework to strengthen the discriminative capability of the multi-scale network. We also plan to integrate thin cloud removal methods to removal the remaining undetected particularly thin cloud. Moreover, since the proposed method cannot be directly applied to cloud shadow detection, we will explore combining a fully supervised cloud shadow network with morphological constraints from cloud–shadow relationships to achieve shadow detection.

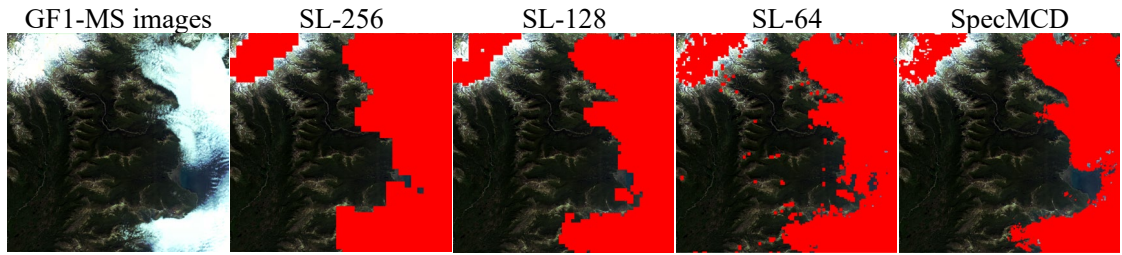

**Fig. 12.** Examples of cloud and cloud shadow detection results obtained by SL-256, SL-128, SL-64, and the SpecMCD method.

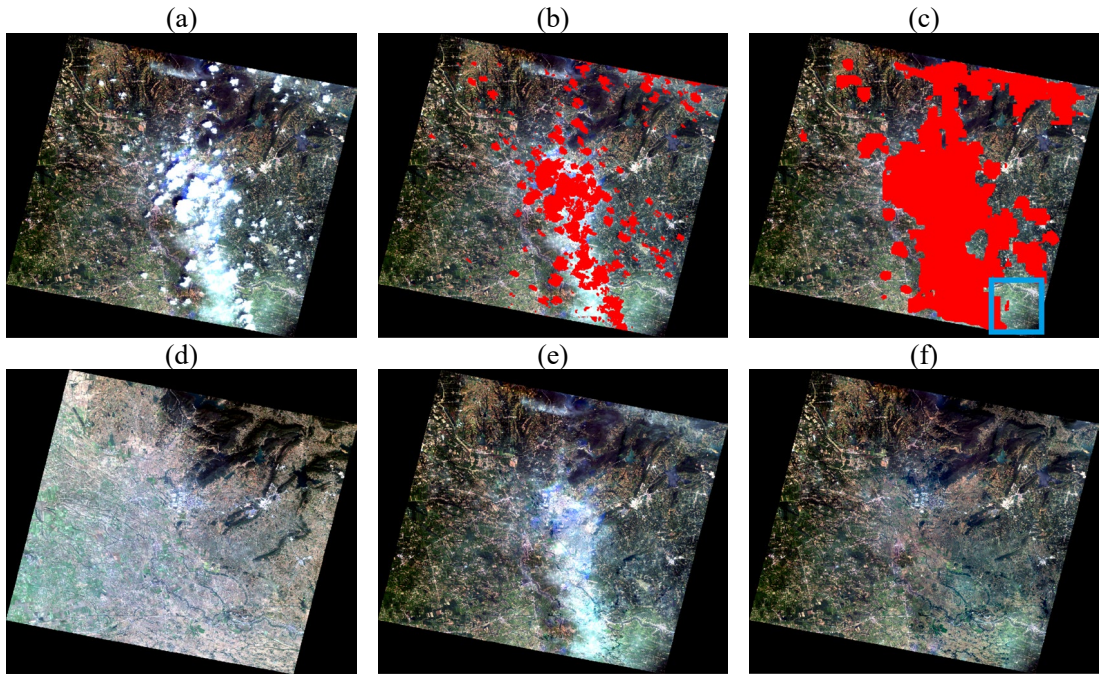

**Fig. 13.** Image reconstruction comparison with different cloud masks. (a) Target image. (b) The RegNetY mask. (c) The SpecMCD mask. (d) Reference image. (e) The RegNetY-based reconstruction. (f) The SpecMCD-based reconstruction.

## 5. Conclusion

In this paper, we proposed a weakly supervised cloud detection method (SpecMCD) that combines spectral features and multi-scale scene-level deep networks. SpecMCD achieves accurate detection of both thick and thin clouds, which significantly reduces omissions occurring in the binary cloud masks. Overall, the proposed method has the following advantages: 1) A progressive training framework is proposed to integrate multi-scale scene-level samples into a single network, which can generate highly accurate multi-scale scene-level cloud probability maps. 2) A differentiated processing strategy is employed to combine multi-scale scene-level network with CTM according to the distribution characteristics of dense and large-area clouds. Furthermore, the probability maps are fused based on the CTM gradient to improve detection performance for dense and large-area clouds. 3) By leveraging differences among multi-scale scene-level masks, adaptive thresholding was extracted to reduce the need for manual threshold adjustment, while distance-weighted optimization further refines the binary masks.

The results of the experiments combining two datasets, WDCD and GF1MS-WHU, demonstrated that SpecMCD improves the F1-score by more than 7.82% compared with other weakly supervised methods, which is effective in reducing omission errors in cloud detection, especially for thin clouds. Nevertheless, the accurate differentiation between clouds and snow remains a challenging task. Future work will explore combining scene-level and pixel-level networks to further enhance detection in dense cloud regions and improve cloud–snow discrimination through a data-driven framework.

## Acknowledgments

This study was supported by the National Natural Science Foundation of China (No. 42130108 and 42101357).

## Disclosure statement

The authors declare that they have no known competing financial interests or personal relationships that could have appeared to influence the work reported in this paper.

## Data availability statement

The data that support the findings of this study are available from the corresponding author, S.Z, upon reasonable request and will be publicly available on GitHub.